\documentclass[conference]{IEEEtran}
\IEEEoverridecommandlockouts
\usepackage{cite}
\usepackage{amsmath,amssymb,amsfonts}
\usepackage{algorithmic}
\usepackage{graphicx}
\usepackage{textcomp}
\usepackage{xcolor}
\usepackage{balance}
\usepackage{colortbl}
\usepackage{multirow}
\usepackage{capt-of}
\usepackage{amssymb}
\usepackage{url}
\usepackage{nameref}
\usepackage{etoolbox}
\usepackage{comment}
\usepackage{enumitem}
\usepackage{tikz}
\usepackage{hyperref}
\def\BibTeX{{\rm B\kern-.05em{\sc i\kern-.025em b}\kern-.08em
    T\kern-.1667em\lower.7ex\hbox{E}\kern-.125emX}}

\def\eg{\emph{e.g., }} 
\def\ie{\emph{i.e., }} 

\definecolor{lightblue}{HTML}{b0dcf4} 
\definecolor{lightgreen}{HTML}{d8ecc4} 
\definecolor{lightyellow}{HTML}{f8f4b0} 
\definecolor{lightturq}{HTML}{b3f2eb} 
\definecolor{DR_orange}{HTML}{ffe3c8} 
\definecolor{DR_blue}{HTML}{c2cae4} 
\definecolor{DR_pink}{HTML}{fecbe5} 
\definecolor{DR_bluegreen}{HTML}{c9ebe7} 
\definecolor{DR_yellowgreen}{HTML}{f6f6cf}

\newcommand{\highlight}[2]{\colorbox{#1}{#2}}

\newcommand{\pquotes}[1]{\textcolor[gray]{0.35}{\textit{#1}}}
\newcommand{\insertfig}{\includegraphics[width=\linewidth]{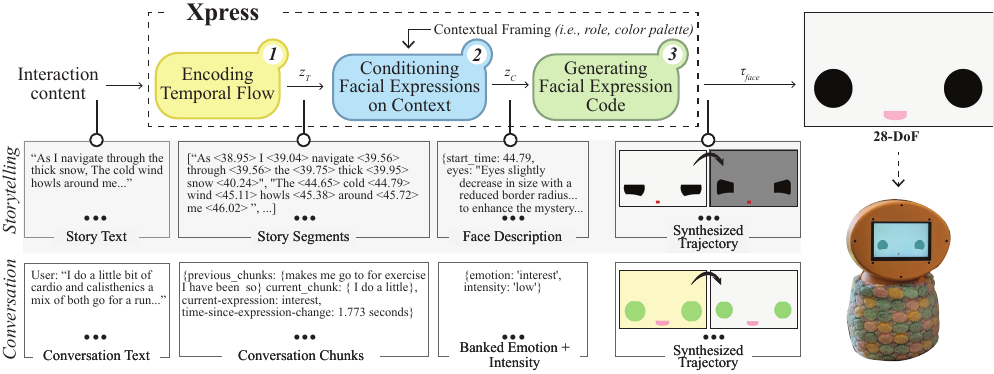}\captionof{figure}{We present Xpress, a system that dynamically generates expressive, context-aware facial expressions for robots. We demonstrate its ability to generate facial expressions in two distinct contexts: storytelling and conversation.}
\label{fig:teaser}}

\makeatletter
\apptocmd{\@maketitle}{\centering\setcounter{figure}{0}\insertfig}{}{}
\makeatother

\begin{document}
\title{Xpress: A System For Dynamic, Context-Aware\\ Robot Facial Expressions using Language Models%

\thanks{\textbf{CRediT author statement}. Conceptualization, Methodology, Writing-Review\&Editing, Visualization (all authors) Formal Analysis, Writing-Original Draft (VNA, MS), Software, Validation, Investigation, Data Curation, Project Administration (VNA) Supervision, Funding (CMH).}
\thanks{\textbf{AI Use.} Text edited with LLM; output checked for correctness by authors.}
}


\author{\IEEEauthorblockN{Victor Nikhil Antony}
\IEEEauthorblockA{
\textit{Johns Hopkins University}\\
Baltimore, MD, USA \\
vantony1@jhu.edu}
\and
\IEEEauthorblockN{Maia Stiber}
\IEEEauthorblockA{
\textit{Johns Hopkins University}\\
Baltimore, MD, USA \\
mstiber@jhu.edu}
\and
\IEEEauthorblockN{Chien-Ming Huang}
\IEEEauthorblockA{
\textit{Johns Hopkins University}\\
Baltimore, MD, USA \\
chienming.huang@jhu.edu}
}


\maketitle

\begin{abstract}
Facial expressions are vital in human communication and significantly influence outcomes in human-robot interaction (HRI), such as likeability, trust, and companionship. However, current methods for generating robotic facial expressions are often labor-intensive, lack adaptability across contexts and platforms, and have limited expressive ranges---leading to repetitive behaviors that reduce interaction quality, particularly in long-term scenarios. We introduce \emph{Xpress}, a system that leverages language models (LMs) to dynamically generate context-aware facial expressions for robots through a three-phase process: encoding temporal flow, conditioning expressions on context, and generating facial expression code. We demonstrated \emph{Xpress} as a proof-of-concept through two user studies ($n=15\times2$) and a case study with children and parents ($n=13$), in storytelling and conversational scenarios to assess the system’s context-awareness, expressiveness, and dynamism. Results demonstrate \emph{Xpress's} ability to dynamically produce expressive and contextually appropriate facial expressions, highlighting its versatility and potential in HRI applications.


\end{abstract}


\section{Introduction}

a Facial expressions are a fundamental aspect of human communication, playing a crucial role in conveying emotions and intentions \cite{buck1972communication, bitti2014expression, birdwhistell2010kinesics}. Similarly, robotic facial expressions are essential not only for expressing emotions and intentions but also for signaling functional and behavioral traits such as intelligence, agency, and personality \cite{broadbent2013robots, hara1995use, diana2011shape}. These expressions significantly influence human-robot interaction outcomes, affecting factors such as likeability, companionship, trustworthiness, and collaboration \cite{zinina2020non, leite2013influence, saunderson2019robots, reyes2016positive}, which are crucial for long-term engagement. Conversely, incongruent robot expressions can negatively impact interactions \cite{bennett2014context, tsiourti2019multimodal, marge2019miscommunication}, underscoring the need for context-appropriate facial expressions.

Two primary approaches exist for producing robot facial expressions: \textit{automated} methods, which leverage machine learning but require extensive training data and often encounter issues with transferability across different contexts and robots \cite{stock2022survey}; and \textit{handcrafting} methods \cite{fitter2016designing}, which are time-consuming and tend to yield repetitive behaviors due to limited emotional diversity. In long-term, repeated interactions, the limited expressive range of existing methods leads to predictable and repetitive behaviors that negatively influence the interaction and relationship \cite{cagiltay2022understanding, tanaka2007socialization}. 

To address the limitations of current robot facial expression generation approaches, we introduce \textbf{Xpress}, a system that leverages language models to process contextual information and dynamically generate robot facial expressions (see Fig.~\ref{fig:teaser}). 

Language models (LMs) demonstrate ability to recognize socio-emotional contexts in text inputs \cite{gong2023lanser, peng2024customising, mishra2023real, gandhi2024understanding, lammerse2022human} and to generate code \cite{liang2023code, karli2024alchemist, perez2021automatic, zhang2023planning}. The core innovation of Xpress is incorporating LMs in a three-phase process to analyze interaction content (\ie robot dialogue, user speech), socio-emotional context, and generation history---enabling dynamic production of contextually appropriate expressions that are expressive.


We validated Xpress as a proof-of-concept through two user studies ($n=15\times2$) and a case study ($n=13$) across two domains. The first user study involved a storytelling robot to test the system’s ability to adapt expressions to complex narrative contexts; a case study with parents and children was also conducted to assess Xpress's validity within this domain. The second focused on a conversational robot to evaluate Xpress's performance in real-time, fluid exchanges. For each user study, we evaluated the system along three key dimensions: how \emph{context-aware} the facial expressions were, how \emph{expressive} they were, how \emph{dynamic} the generations were.

This work makes three key contributions: 
\begin{enumerate}[leftmargin=*]
\item Design, development, and open-sourcing of \textbf{Xpress}.
\item Demonstration of \textbf{Xpress} in two domains---\emph{storytelling} and \emph{conversational interaction}---as a proof-of-concept.
\item Insights from our exploration with \textbf{Xpress}, highlighting the potential and challenges of using state-of-the-art LMs for generating facial expressions in robotic systems.
\end{enumerate}


\section{Background and Related Work}

\subsection{Emotions and Human-Robot Interaction} 

Robots that express emotions enhance social presence and user engagement, making interactions more meaningful \cite{breazeal2003emotion, hara1995use, leite2008emotional}. Emotion expression improves perceptions of likeability, companionship, trust, intelligence, and collaboration \cite{wiese2017robots, broadbent2013robots, zinina2020non, leite2013influence, saunderson2019robots, reyes2016positive}, and fosters human empathy \cite{kim2009can}. In therapy \cite{shibata2011robot} and education \cite{saerbeck2010expressive}, emotion expression enhances patient experiences and motivates students, demonstrating their practical value. However, robots must express emotions appropriately, as mismatched responses can harm interactions \cite{bennett2014context, tsiourti2019multimodal}. To ensure accuracy, current emotion recognition methods rely on cues such as facial expressions and body language \cite{el2005real, kleinsmith2011automatic} assuming strong correlations between physical expressions, subjective feelings, and social meaning. These methods often rely on discretizing emotions into preset categories~\cite{ekman1992facial}, which can fail to capture the nuances of complex interactions. Emotions in social contexts are rarely discrete, making rigid categorization limiting~\cite{jung2017affective, scarantino2012define, greenaway2018context}.

Advancements in LMs have shown state-of-the-art potential in recognizing socio-emotional context across various domains \cite{peng2024customising, gandhi2024understanding}; LMs can interpret nuanced emotion in interviews with children \cite{lammerse2022human}, dialogic interactions \cite{mishra2023real}, and speech data \cite{gong2023lanser}. \textbf{Xpress} leverages this capability to understand socio-emotional context to capture different contextual variables to inform the expression generation.

\subsection{Generation of Facial Expressions for Robots} Facial expressions are a common method for affect communication in robots \cite{fitter2016designing, kalegina2018characterizing}. Generating appropriate facial expressions has been a focus of research, with methods ranging from handcrafted designs \cite{fitter2016designing, herdel2021drone}---which are labor-intensive and limited in expressiveness—to automated methods using pre-defined scripts \cite{acosta2008robot} or machine learning techniques (\eg k-nearest neighbors \cite{trovato2012development} and generative adversarial networks \cite{rawal2022exgennet}). These automated methods often require large datasets and lack transferability across robotic systems, especially those with varying levels of anthropomorphism.

Language models have demonstrated state-of-the-art capabilities in generating programs for robots across different platforms and tasks, such as manipulation and planning \cite{liang2023code, karli2024alchemist, perez2021automatic, zhang2023planning}. Recent work has also utilized LMs to generate individual non-verbal behaviors for robots \cite{mahadevan2024generative}. \textbf{Xpress} integrates LM code generation capability and pushes the state-of-the-art to generate robotic facial behaviors across complex interactions.


\section{Xpress: Generating Dynamic, Context-Aware Facial Expressions}

\textbf{System Overview.} \textit{Xpress} follows a three-phase process to generate context-aware facial expressions as a temporal trajectory, denoted as $\tau_{\text{face}}$, which is conditioned on both the interaction content (\eg robot speech, user speech) and socio-emotional context (\eg tone, interaction background). Fig.~\ref{fig:teaser} illustrates Xpress along with example inputs and generations.

\textit{\highlight{lightyellow}{Phase 1} Encoding Temporal Flow.} Xpress, first, processes the interaction content to capture the temporal dependencies (\eg timing, flow) of the interaction content and to encode the temporal information into the interaction content to create an intermediate representation $z_T$. 

\textit{\highlight{lightblue}{Phase 2} Conditioning Facial Expressions on Context.} Xpress analyzes $z_T$ along with contextual framing information (\eg role, previous interaction) to capture socio-emotional context (\eg emotional tone, expected reaction) and encodes it as a high-level abstraction $z_C$ that defines the robot face, enabling the representation of a spectrum of emotions rather than restricting them to discrete categories.

\textit{\highlight{lightgreen}{Phase 3} Synthesizing the Trajectory.} Finally, the abstract representation $z_C$ is used to sequentially synthesize facial expressions to populate the the expression trajectory $\tau_{\text{face}}$; this trajectory can be then executed by a robot to produce context-appropriate expressions for the given input.

\textbf{Xpress in action.} To demonstrate and validate the capabilities of Xpress, we implemented it in two distinct domains: \textit{Storytelling} and \textit{Conversational Interaction}. We selected storytelling because it presents complex temporal and socio-emotional challenges, each story requires tailored expressions to ensure contextual appropriateness. 
We chose conversation to demonstrate its effectiveness in handling real-time, fluid interactions. For these two domains, we use Xpress to generate facial expressions for a custom 28-DoF animated face, created to support a wide range of expressions while capturing the abstract nature of emotional communication (see Fig. \ref{fig:teaser}). 



\section{Face System}





Our custom face system\footnote{adapted from: \url{https://github.com/mjyc/tablet-robot-face}} is designed for flexibility and a high degree of manipulability, providing complete control over a wide range of expressions. It uses anime.js~\footnote{\url{https://animejs.com/documentation/}} to allow for dynamic transitions, with adjustable animation durations and easing effects. Given the semi-abstract nature of the face, the large number of possible manipulations (28-DoF) (see Appendix I), and the animation complexities, this face system poses a challenge for face generation (Fig.~\ref{fig:teaser}).


\subsection{Code Generation using Language Model} \label{sec:code-gen-model}
 Xpress's \highlight{lightgreen}{Phase 3} must be able to generate the code to render the face as described by $z_C$ to use this system. To achieve this functionality and animate our robot face to produce a wide expressive range, we prompt a GPT-4 model to generate executable code tailored for our face system. 
The code generation prompt instructs the model to assume the role of a ``creative and talented JavaScript programmer,'' tasked with generating JS code that animates the robot’s face for the given face description, $z_C$. See Appendix IIA for full prompt. 
\begin{itemize} [leftmargin=*]
    \item \textit{Functional Overview.} The prompt details the capabilities and limitations of facial elements (eyes, eyelids, mouth, and the face background), specifies control parameters for each element (\eg scaling, positioning, rotation, and color adjustment), and provides corresponding example code snippets. Avoiding restrictions to a high-level API encourages code generation that aligns closely with the specifications in $z_C$. 
    \item \textit{Rules.} A set of rules are given to define standards for the generated code. Additional guidelines are provided to encourage smooth animation and avoid abrupt transitions.
    \item \textit{Steps.} The model is instructed to follow a step-by-step process. Step 1: Generate an initial version of the code based on the face description in $z_C$. Step 2: Verify that all requested changes are reflected accurately and that no unnecessary residuals from the prior face state exist. Step 3: Check that the code adheres to the specified rules and refine the output if necessary. These steps ensure a structured workflow, enabling the model to focus on accuracy. 
\end{itemize}

\begin{figure*}
\centering
\includegraphics[width=\textwidth]{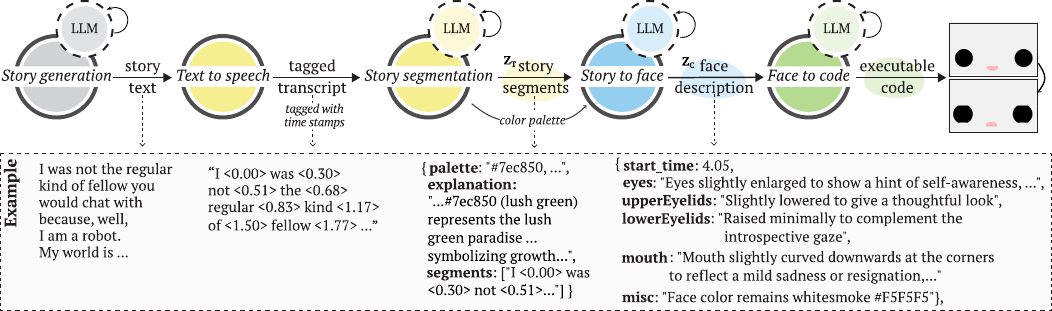}
\caption{Xpress pipeline for storytelling content generation.}
\label{fig:storytelling-pipeline}
\end{figure*}

\section{Storytelling System}

We developed a storytelling system to showcase Xpress'\footnote{\url{https://github.com/vantony1/Xpress}} capability in generating context-aware facial expressions over varied socio-emotional context. Fig.~\ref{fig:storytelling-pipeline} shows the end-to-end overview of the storytelling system\footnote{\href{https://tinyurl.com/bd4pc9dn}{Generation Samples} and \href{https://github.com/intuitivecomputing/Publications/blob/e4ebaa0c0d96d9c6e1bbc2de69609c0cee792feb/2025/HRI/Supplementary_2025_HRI_Antony_Xpress.pdf}{Appendix with Supplementary Materials}
}.



\subsection{Using Xpress for Storytelling Face Generation}

\textit{\highlight{lightyellow}{Phase 1}} 
To capture the temporal flow of the story, we produce a word-level transcript with timepstamps for a given story. We synthesize the robot's speech using a Google text-to-speech model, transcribe it with a Google speech-to-text model, and process the timestamped transcript using GPT-4. This LM is prompted to adopt the persona of an \textit{``expert, creative, and talented animator working at Disney’’} on a robot storytelling project (Appendix IIC contains the full prompt).

\textit{Story Segmentation.} The LM segments the timestamped transcript into discrete chunks based on significant shifts in the narrative or emotional tone. These segments inform when the robot should alter its facial expressions to reflect preemptive, timely, or delayed reactions, aligning with the unfolding story.

\textit{Color Palette Generation.} Alongside segmentation, the LM generates a color palette to complement each story segment's emotional tone. Each color is meant to enhance the robot’s expressiveness and has an explanation detailing how it corresponds to specific emotions or narrative moments.

The segmented and timestamped story transcript forms the temporal embedding, $z_T$, capturing both the temporal flow and emotional shifts throughout the narrative. The color palette is passed to Phase 2 as a meta-context variable.

\textit{\highlight{lightblue}{Phase 2}} 
 Another GPT-4 LM processes each story segment $z_T[i]$ sequentially and generates detailed descriptions of facial expressions for the robot.
This LM is prompted (see Appendix IID) to adopt the same persona as from \textit{Phase 1} and is tasked with designing facial expressions for a robot storyteller.

\textit{Face Specification.} The model is provided a complete description of the robot’s face, detailing individual elements and capabilities such as eye movements and eyelid positioning. The prompt explains how each element can be manipulated, to give the model a understanding of how to create facial expressions.

\textit{Task Description.} The model is instructed to utilize the described capabilities to creatively design expressive facial expressions tailored to the emotional tone of each story segment. The goal is to ensure that the robot’s expressions are engaging and contextually appropriate.

\textit{Workflow Steps.} The model follows a structured workflow:

\begin{itemize} [leftmargin=*]
    \item \textit{Understanding the Segment:} For the given segment $z_T[i]$, the model analyzes the content and emotional context, considering the narrative progression up to that point.
    \item \textit{Designing the Expression:} The model designs an appropriate facial expression by specifying adjustments to the facial elements and its timing within the segment to align the expression with the story's emotional tone and context.
    \item \textit{Coherence Check:} The model checks the expression for consistency with prior ones and for use of the face's abilities.
    \item  \textit{Revision and Output:} If necessary, the model revises the expression description to better fit the narrative and adhere to the given guidelines. 
\end{itemize}



\noindent The generated face descriptions form the high-level abstraction, $z_C$. 
Processing $z_T$ sequentially conditions the facial expression at time step $i$, $z_C[i]$ on the current segment ($z_T[i]$), on the preceding segments ($z_T[0, ..., i-1]$) and the previous expression descriptions ($z_C[0, ..., i-1]$). This sequential processing promotes coherent transitions between expressions, as past expressions and contexts influence future ones, to enable a smooth emotional progression across the story.

\textit{\highlight{lightgreen}{Phase 3}} Facial expressions, $z_C$, are sequentially processed by the code generation LM (\textsection \ref{sec:code-gen-model}) to produce face programs. This results in a timestamped trajectory of facial expressions, $\tau_{\text{face}}$, which is synchronized with the story’s delivery.




\subsection{Story Delivery}
To overcome the computational overhead ($\sim$7 minutes for a 500-word story), we pre-generate facial expressions using Xpress. The story is stored in a JSON object with base64-encoded audio and timestamps linked to JavaScript code for each expression. During storytelling, the audio plays, and a Node.js server triggers facial expressions based on the timestamps and animates the mouth to simulate speaking.

\section{Proof-of-Concept: Storytelling} \label{sec:val-storytelling}

\begin{figure*}[t]
\centering
\includegraphics[width=\textwidth]{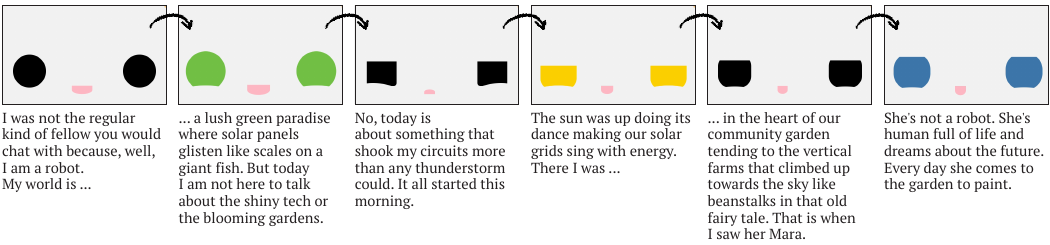}
\caption{Example delivery of story generated using Xpress showing robot faces and the corresponding story text.}
\label{fig:example-storytelling}
\end{figure*}

To validate Xpress's ability to generate dynamic, context-aware facial expressions, we conducted a study where participants evaluated a robot telling 12 children's stories across three genres: \textit{solarpunk}, \textit{horror}, and \textit{space adventure} (each $\sim$500 words).  Fig.~\ref{fig:example-storytelling} shows a story with corresponding expressions.  

\subsection{Study Design} 
\subsubsection{Procedure} The robot narrated three stories to each participant, with one story randomly selected from each of three genres, and a 10-second break between stories. Participants completed a 5-point Likert-scale survey assessing their perception of the robot’s facial expressions in terms of context-appropriateness, timeliness, expressiveness, and repetitiveness. The experimenter then conducted a semi-structured interview to understand participants’ overall experience and perceptions.

\subsubsection{Participants} We recruited 18 participants (8M/10F, ages 18-39, $M=25.3,SD=5.5$) through a community newsletter. The study took roughly 25 minutes and participants provided informed consent and were compensated at \$15/hr as approved by our institutional review board (IRB). We excluded two participants for failing to follow study instructions and one participant due to a node.js server malfunction.

\subsubsection{Generation Metrics} We computed the \textit{Executable Code Percentage} \ie the percentage of generated facial expression code that runs without errors. We also noted the \textit{Number of Unique Faces} generated across the 12 stories. 

\subsubsection{Perception Metrics} We created two scales: \textit{Context Alignment Score} (seven-item scale (Cronbach's $\alpha = 0.87$)) evaluates the degree to which the robot's facial expressions aligned with the context of the story and the \textit{Expressiveness Score} (four-item scale (Cronbach's $\alpha = 0.76$)) assesses the expressive range and clarity of the robot's facial expression. See Appendix III for a breakdown of the subjective scales. Lastly, \textit{Context Match Percentage} was derived from an interview question (``How often was the face appropriate for the story context? Provide a number between 0\% and 100\%.'').

\begin{figure}[h]
\includegraphics[width = \columnwidth]{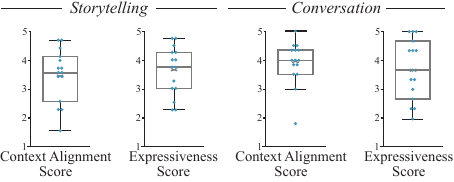}
\caption{Participants' perception of storytelling and conversational systems' faces. Cross is mean; box shows quartiles.}
\label{fig:findings}
\end{figure}

\subsection{Findings} \label{story-val-findings}
The quantitative results are presented in Fig. \ref{fig:findings}-storytelling and we detail the key observations from the post-study interviews and the survey data analysis to ground our findings in the nuances of the participants' perceptions and experience. We refer to participants along with their respective quantitative metrics scores as follows: \textbf{PID} (\textit{Context Match Percentage/Context Alignment Score/Expressiveness Score}).

\subsubsection{Dynamic Generation (Generation Performance Metrics)}
Across the 12 stories, the Xpress system successfully generated 245 unique robot facial expressions. Each participant experienced approximately 60 distinct facial expressions during their interaction. Notably, 100\% of the facial expression code generated by Xpress was executable without errors.

\subsubsection{Context-Awareness}
Participants generally found the robot's facial expressions to align well with the context of the story, as reflected by the \textit{Context Match Percentage} averaging 71.67\% (SD = 21.87\%). Additionally, the \textit{Context Alignment Score} had a mean score of 3.48 (SD = 0.93), which is above neutral ($t(14) = 1.98, p = .034$). 


Most participants stated that the expressions typically matched the expected emotions except for occasional discrepancies. For instance, A10(72.5\%/3.71/4.5) shared: \pquotes{``They were pretty clear at conveying what I think is the expected emotion for the most part ... I think just once in a while, they seemed a little bit off, but for the most part they helped carry the story''}. 

A5(95\%/4.71/3.25) highlighted the synchronization between the robot's facial expressions and the story's events, stating: \pquotes{``The timing of [the facial expressions] and the rhythm was really interesting to see that a robot could do that.''} A4 (90\%/4.43/3.00) also noted that the expression timing and interpretation lag with the robot was similar to that during an interaction with a human: \pquotes{``I'd have to basically listen for one or two more sentences ahead to fully understand what [the facial expressions] were trying to convey. That's kind of like with a normal person, too.''}

Participants cited the intensity of the emotion as a reason for perceiving expressions as mismatched. A12(87.5\%/2.57/4.25) shared: \pquotes{``Sometimes I felt that it didn't match the emotions of the story, but like most of the time, it was spot on... [recalling one specific scene] the robot was supposed to be like a little bit scared, but it didn't look *that* scared...''}. This suggests that while the robot conveyed the correct emotion, the intensity did not always match participants' expectations given the context.

\subsubsection{Expressiveness} The robot's facial expressiveness was rated positively with an average \textit{Expressiveness Score} of 3.62 (SD = 0.86), above neutral ($t(14) = 2.78, p = .007$). Participants frequently commented on the diversity and range of the expressions. A11 (35\%/2.29/2.25), for example, appreciated the variety of expressions, noting that the robot avoided repetitive gestures: \pquotes{``the diversity of its expressions was very nice. I feel like I didn't see too many repeats... It could show, I think, a wide range of emotion.''}. Similarly, A6(95\%/4.71/4.75) noted the robot's ability to portray nuanced and subtle emotions: \pquotes{``there was different degrees of expression ...it gave the robot the ability to convey more subtle emotions...''}

However, not all participants found the robot's expressiveness intuitive. Two participants struggled to associate the robot's changing facial features---specifically the shifting shapes and colors---with emotions. A8 (20\%/2.28/2.5) commented: \pquotes{``I think the color changing didn't really relay anything... shapes aren't emotion...Like I didn't correlate that with emotion intuitively.''}; This challenge negatively affected their perception of the robot's emotional range and the appropriateness of its expressions. 

\begin{figure*}
\centering
\includegraphics[width=\textwidth]{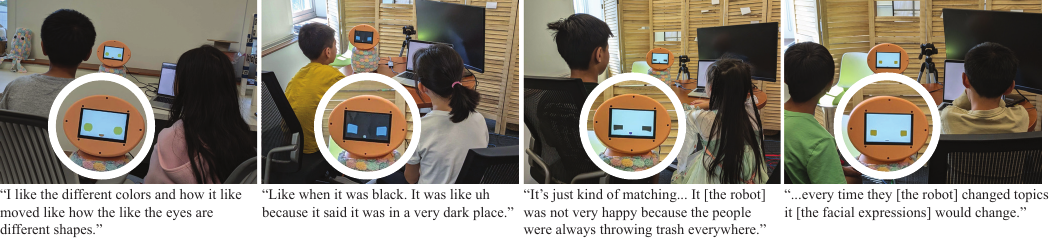}
\caption{Children watched and evaluated our robot narrating stories generated using Xpress.}
\label{fig:case-study}
\end{figure*}

\section{Case Study: Storytelling with Children}
Storytelling robots have been proposed to foster cognitive and social growth in children \cite{leite2015emotional, antunes2022inclusive}. To evaluate Xpress in this context, we conducted a case study with children, assessing its performance within the target population.


\subsection{Study Design}
After receiving informed consent from parents and assent from children, the experimenter introduced the robot and its role as a storyteller. The experimenter turned on the system, which introduced itself and told a story. After the first story, children were given the option to either continue or stop listening. The system narrated \textit{space adventure} and \textit{solarpunk} stories (\textsection \ref{sec:val-storytelling}). Once the children chose to end the session, they were asked to answer three questions using a 5-point scale response in the form of smiley faces and clear labels to assess how much they liked the story and facial expressions, and whether they would like to listen to more stories. We then interviewed the children about their experience and the parents about their perception of the system and its utility.

\textbf{Participants.} We recruited children ($n = 8$, 5M/3F, ages 7-11) and parents ($n = 5$) using a community group chat. Each study took about 20 minutes and was approved by our IRB. We compensated the children with robot stickers and candy.


\subsection{Findings} 

The children expressed a high level of enjoyment in their interaction with the robot. They rated both the robot's storytelling abilities ($M = 4.13, SD = 0.64$) and facial expressions ($M = 4.38, SD = 0.74$) positively; their positive experiences were also reflected in their willingness to listen to additional stories ($M = 3.63, SD = 0.92$). Fig.~\ref{fig:case-study}  presents quotes from the children alongside generated facial expressions.

\subsubsection{Perception of Facial Expressions and Appearance}

Children responded positively to the robot's facial expressions. 
C7 expressed that she liked \pquotes{``the different colors and how it moved, like how the eyes are different shapes.''} Additionally, children were able to connect the robot's changing facial expressions to the content of the stories. For example, C5 observed, \pquotes{``sometimes [robot] was not very happy because the people were always throwing trash everywhere.''} indicating that the facial expressions were reflective of the story content.

\subsubsection{Story Preference and Possible Extensions} Overall, children enjoyed the stories, with each child listening to an average of two stories (range: 1–3); however, they desired a greater variety in both story genre and duration. Similarly, parents highlighted the need to adjust story complexity based on the child’s age. Children envisioned using the robot for bedtime stories or to alleviate boredom, while parents imagined their children creating stories with the robot. Moreover, children wanted more interactivity, with options to tell the robot stories or have it ask for their story preferences. Parents noted that incorporating interactions during storytelling sessions could help maintain children’s engagement during longer interactions.

\begin{figure}[b]
\centering
\includegraphics[width=\columnwidth]{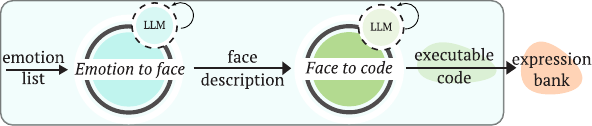}
\caption{Pipeline for pre-generation of expression bank.}
\label{fig:pre-generation}
\end{figure}

\section{Conversational System} 
To demonstrate Xpress's ability to generate expressive, context-aware facial expressions in other interaction paradigms, we adapted it for real-time conversational interactions, overcoming computational overhead.

\begin{figure*}
\centering
\includegraphics[width=\textwidth]{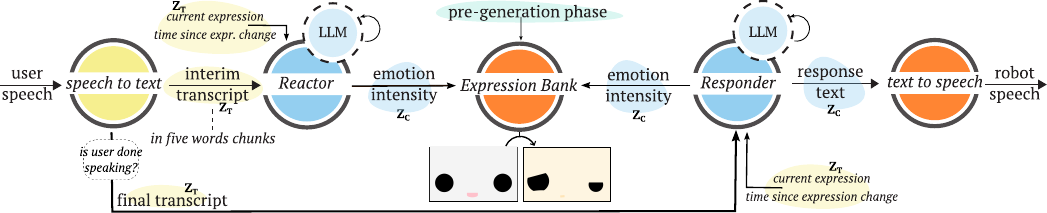}
\caption{Xpress pipeline for real-time interaction using a pre-generated expression bank.}
\label{fig:conversation-pipeline}
\end{figure*}


\subsection{Using Xpress in Real-time}

\textit{Adapting Xpress.} We modified two key aspects of \textit{Phase 2} and \textit{Phase 3} to enable robots to understand context and execute dynamic, appropriate facial expressions in real-time. Specifically, we simplify the face description generation by outputting $z_C$ as discrete emotion-intensity pairings in \textit{Phase 2} and pre-generate an expression bank for \textit{Phase 3}---lowering computation lag by pre-processing the most time-consuming step of Xpress (face description and code generation). Fig.~\ref{fig:conversation-pipeline} provides an overview of the conversational system.

\textit{\highlight{lightturq}{Pre-generation (Phase 2+3)}} We generated a facial expression bank based on predefined emotions: \textit{happy, sad, surprise, stress, calm, confusion, tired, interest, concern, fear, disgust, angry, neutral}~\cite{russell1980circumplex}. Each emotion has two intensity levels: low and high. An LM (Appendix IIE) describes facial expressions for each emotion and intensity, prompted similar to the Phase 2 face description LM in the storytelling system (see Fig.~\ref{fig:pre-generation}).


\textit{Workflow Steps.}
The model follows a structured workflow:

\begin{itemize} [leftmargin=*]
\item \textit{Color Palette Selection:} The model first selects a color palette using color theory that complements the emotion.
\item \textit{Facial Element Adjustment:} The model describes an appropriate facial expression by adjusting facial elements to reflect the emotion.
\item \textit{Expression Variation:} The model generates two variants of the expression description---for low and high intensity.
\item \textit{Revision and Output:} The model checks the descriptions for coherence, making revisions if necessary.
\end{itemize}


The resulting descriptions form the higher-level representation $z_C$, which is used to pre-generate the emotion-expression bank via the code generation model. This bank allows the system to render expressions during real-time interaction.


\highlight{lightyellow}{Phase 1} While the \textit{robot is listening}, the user's speech is transcribed in real-time and segmented into 5-word chunks. These, along with time since the last facial change and current expression, form the temporal encoding $z_T$. Once the \textit{user finishes}, the full transcription replaces the partial transcripts in $z_T$, supporting temporal coherence across the conversation.

\highlight{lightblue}{Phase 2} As the \textit{robot listens}, a GPT-4o-mini LM (Fig.~\ref{fig:conversation-pipeline}-Reactor; Appendix IIF for prompt) receives $z_T$ and acts as an \textit{``empathetic and supportive social robot''} to make the user feel heard. The LM evaluates $z_T$ to get context and react with an appropriate emotion and intensity from the expression bank---or opts for no change to avoid frequent expression shifts.

Once the \textit{user finishes speaking}, to generate the robot's response, a GPT-4o LM is prompted (Fig.~\ref{fig:conversation-pipeline}-Responder; Appendix IIG for prompt) to be an \textit{``empathetic and supportive social robot''} to facilitate dialogue around questions that promote reflection on the user’s physical and emotional state. The LM follows these steps:
\begin{itemize} [leftmargin=*]
\item \textit{Generate Verbal Reply:} The model responds to the user’s speech by asking the next question or a follow-up to encourage reflection.
\item \textit{Select Appropriate Emotion for Response:} The LM also selects an emotion-intensity dyad $z_C$ for its own response, considering the user's response, its response, current expression, and time since expression change.
\end{itemize}

\noindent This process generates the emotion-intensity dyad $z_C[i]$ for timestep $i$, conditioned on previous expressions $z_C[0, …, i-1]$ and interaction content $z_T[0, …, i]$.

\highlight{lightgreen}{Phase 3} While the \textit{robot is listening}, the emotion-intensity dyad from $z_C$ is displayed using the emotion bank in real time. While \textit{robot is responding}, the selected expression in $z_C$ is displayed for the first 3 seconds of the robot's speech before resetting to a neutral face for the remainder of the response.

\begin{figure*}
\centering
\includegraphics[width=\textwidth]{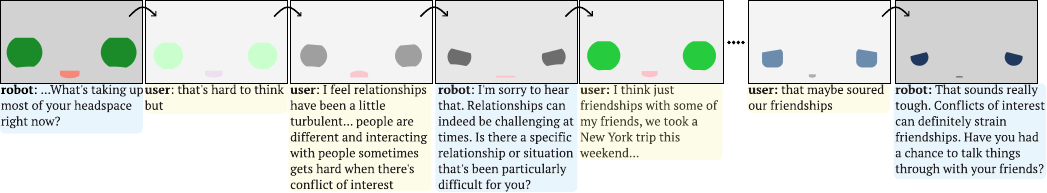}
\caption{A snippet of the conversation between a participant and the conversational robot showcasing the robot's facial reactions.}
\label{fig:conversation-example}
\end{figure*}

\section{Proof-of-Concept: Real-time Conversation}
To validate Xpress in real-time interactions, we conducted a study with a robot performing a well-being check-ins with users through a set of questions. Three expression banks were generated, with one randomly selected for each interaction. Fig.~\ref{fig:conversation-example} illustrates a conversation and corresponding expressions.

\subsection{Study Design}
\subsubsection{Procedure} Participants conversed with the robot as it asked them the daily check-in questions followed by responses and follow-up questions. After the conversation, they completed a 5-point Likert-scale survey assessing the facial expressions' appropriateness, timeliness, expressiveness, and repetitiveness. A semi-structured interview was then conducted to explore their overall experience and preferences.

\subsubsection{Participants} We recruited 15 participants (6M/9F, ages 19-23, $M=21.1,SD=1.51$) through a community newsletter and mailing lists. The study took roughly 20 minutes. Participants provided informed consent and were compensated at \$15/hr as approved by our IRB.
\subsubsection{Generation Metrics} We computed the \textit{Executable Code Percentage} \ie the percentage of generated facial expression code that runs without errors. We also note the \textit{Number of Unique Faces} generated across the 3 banks. 

\subsubsection{Perception Metrics} We created two scales: \textit{Context Alignment Score} (six-item scale (Cronbach's $\alpha = 0.81$)) evaluates the degree to which the robot's facial expressions aligned with the conversation context and \textit{Expressiveness Score} (three-item scale (Cronbach's $\alpha = 0.84$)) assesses the expressive range and clarity of the robot's facial expressions. See Appendix IV for subjective scales breakdown. Lastly, \textit{Context Match Percentage} was derived from an interview question (``How often was the face appropriate for the conversation context? provide a number between 0\% and 100\%.'').

\subsection{Findings}

The quantitative results are presented in Fig.~\ref{fig:findings}-conversation and we details the key observations from the post-study interviews and the survey data analysis to ground the results in the nuances of the participants' perceptions and experience. We refer to participants along with their respective
quantitative metrics scores as follows: \textbf{PID} (\textit{Context Match Percentage/Context Alignment Score/ Expressiveness Score}).


\subsubsection{Dynamic Generation}
Across the 3 generated banks, Xpress produced 72 faces, with 100\% of the generated code executing without errors. There was no statistically significant variance in the banks’ impact on the quantitative results.

\subsubsection{Context-Awareness} 
Participants generally perceived the robot's facial expressions to align well with the conversational context. This is supported by the \textit{Context Match Percentage} averaging 75.00\% (SD = 23.10\%) and the \textit{Context Alignment Scale} which had a mean score of 3.90 (SD = 0.75), above neutral, $t(14) = 4.63, p < .001$. 


Most participants felt that the robot's expressions appropriately matched the content of the conversation. For instance, B10 (98\%/5/5) expressed that the interaction \pquotes{``almost felt like talking to therapist or a friend...it's actually like trying to understand what I'm saying and the facial expressions are according to that.''} However, participants noted issues with the intensity of expressions, which were not always proportional to the situation. B7 (60\%/3.5/5) commented, \pquotes{``It showed a lot more emotions than it needed to, like there was a lot of emotion for 'what did I eat?'... it did match the emotion of what it was talking about... but then there was extra emotion''}.

Expression timing influenced perception of context awareness with some participants noting a lag. For example, B8 (80\%/5.0/4.5) responded to the \textit{Context
Match Percentage} question saying,\pquotes{``I'd say 80\%. And that 20\% is the lagging between reaction and the current topic, it doesn't react as fast as human.''}. Other participants observed that the robot's responses were overly timely; B4 (70\%/3.83/2.33) shared \pquotes{``if [the robot] thinks that I'm sleeping a little bit late, suddenly the face feel like a frowning face and then I said that I'm feeling energetic, then [the robot's] face will become a smiling face. So it's very like timely response but it's a bit too timely.''}

\subsubsection{Expressiveness} 
The robot's facial expressiveness was rated positively with an average of 3.69 ($SD = 1.06$), above neutral ($t(14) = 2.52, p = .01$). While participants observed various facial expressions during the interaction, they noted that those expressions appeared to be heavily influenced by the conversation's emotional tone. For instance, in predominantly positive conversations, only positive expressions were experienced. Some participants did note a limited depth in the expressions, leading to mismatches in intensity or emotional appropriateness. B1 (100\%/4.5/3.33) commented, \pquotes{``I feel like there's a limited set of emotions ... if I'm talking about something depressing, it would feel sad but not empathetic.''}


\section{Discussion}
Overall, we find that \textbf{Xpress} can dynamically generate expressive and contextually appropriate robot facial expressions across two versatile and different interaction contexts.

\textbf{Catalyst for Humanizing HRI.} Participants’ experiences emphasized how \textbf{Xpress}-generated expressions enhance human-robot interaction, extending beyond mere emotion communication. Notably, the facial expressions had a humanizing effect on the perception of the robot with A3(90\%/4.00/3.75) sharing that the interaction \pquotes{``didn't feel like a robot was telling me a story. It fell closer to a human being telling me a story.''} Similarly, A7(80\%/4.14/3.75) remarked that \pquotes{``[the robot's] facial expressions had a very humanizing quality... from the beginning to the end, I definitely had a difference of perception. It was a more human perception and that was due to the facial expressions.''}. 

During conversational interactions, the reactivity and the alignment of the facial expressions made the participants feel heard with B4(70\%/3.83/2.33) noting that \pquotes{``these reactions will actually like, kind of make me feel more willing to share more... [the robot] having reaction made me feel more reassured that it is actually listening.''} These experiences highlight the value of context-aware expression generation in social robots and mark Xpress as a step beyond categorical emotion approaches \cite{mishra2023real}, aligning with contemporary emotion theory \cite{jung2017affective, scarantino2012define, greenaway2018context}.

Xpress achieves $\sim$70\% Context Match Percentage, demonstrating a robust proof of concept aligned with benchmarks of human understanding of brief expressive behaviors \cite{ambady1992thin}, while individual differences in perception highlight the need for further research with controlled baselines.

 
\textbf{Co-Creation of Robot Behaviors.} Xpress has demonstrated a capability to augment the behavior generation pipelines for robots adding to the evidence on the utility of language models for content generation~\cite{geyer2021hai, antony2024id, louie2020novice, lopez2024co, lawton2023drawing}. However, bias in the language models could manifest in inappropriate or harmful robot behaviors~\cite{ayyamperumal2024current}; addressing this risk is critical for ethical deployment of such systems in sensitive contexts like mental health support or when interacting with vulnerable populations, such as children.

Adapting Xpress as a co-creation system with a human-in-the-loop approach could mitigate these risks while simplifying the existing manual yet
laborious process of robot expression generation \cite{dennler2024pylips, alves2022flex, schoen2023lively}. By enabling stakeholders, such as educators and caregivers, to co-create content and robot behaviors tailored to their specific needs~\cite{hatch2023pathwise}, the system can avoid unsuitable outputs. Additionally, co-created behaviors can inform and condition LM outputs, achieving auto-generated behaviors that better align with user safety and social norms.


\section{Limitations \& Future Work}

Human communication is inherently multi-modal, seamlessly integrating verbal and non-verbal cues~\cite{argyle2013bodily}. However, \textbf{Xpress} processes only textual inputs to generate context-aware but uni-modal behavior \ie facial expressions. Social interactions encompass more than just words; they involve vocal prosody, body language (\eg posture, gaze), environmental cues, and cultural context \cite{mehrabian2017nonverbal, matsumoto2009cross}. Analyzing prosodic features such as intonation and tempo~\cite{schuller2013interspeech, cowie2001emotion}, along with facial expressions~\cite{stiber2022modeling}, could provide Xpress with additional information, enhancing its ability to understand socio-emotional context and generate more nuanced behaviors.

As speech tone, body posture, and facial expressions all shape the socio-emotional context, multi-modal behaviors are essential for effective robot communication \ie facial expressions must be synchronized with physical movements, speech tone, and pacing~\cite{tsiourti2019multimodal}. Multi-modal behaviors not only broaden a robot's expressive range but also sustain user engagement over time. Our case study showed that children wanted the robot to move during storytelling, while the conversational validation highlighted the need for more subtle emotions achievable through multi-modality. By extending \textit{Phase 2} and \textit{3}, \textbf{Xpress} could generate multi-modal robot behaviors.

A key limitation of Xpress is generation time, requiring adaptation for real-time performance. Leveraging smaller language models could enable real-time generation without a pre-generation phase (Fig. \ref{fig:pre-generation}).

\section*{Acknowledgment}
This work was supported in part by the JHU Malone Center for Engineering in Healthcare.

\clearpage
\balance

\bibliographystyle{IEEEtran}
\bibliography{IEEEabrv,IEEEexample}

\end{document}


\title{Xpress: A System For Dynamic, Context-Aware Robot Facial Expressions using Language Models:\\
Supplementary Materials
}

\author{\IEEEauthorblockN{Victor Nikhil Antony}
\IEEEauthorblockA{
\textit{Johns Hopkins University}\\
Baltimore, MD, USA \\
vantony1@jhu.edu}
\and
\IEEEauthorblockN{Maia Stiber}
\IEEEauthorblockA{
\textit{Johns Hopkins University}\\
Baltimore, MD, USA \\
mstiber@jhu.edu}
\and
\IEEEauthorblockN{Chien-Ming Huang}
\IEEEauthorblockA{
\textit{Johns Hopkins University}\\
Baltimore, MD, USA \\
chienming.huang@jhu.edu}
}

\maketitle

\section{Appendix: Face System}

\begin{table}[h]
\caption{Face System Degrees of Freedom}
\label{tab:face-DoF}
\resizebox{\columnwidth}{!}{%
\begin{tabular}{lcp{4.5cm}}
Parts & \multicolumn{1}{l}{DoF} & \multicolumn{1}{l}{Description} \\ \hline
\multicolumn{1}{l|}{Background} & 1 & Color \\
\multicolumn{1}{l|}{Left/Right Upper/Lower Eyelid} & 8 & Position (Y), Rotation \\
\multicolumn{1}{l|}{Left/Right Eye} & 12 & Color, Position (X, Y), Border Radius, Size (Height, Width) \\
\multicolumn{1}{l|}{Mouth} & 7 & Color, Position (X, Y), Border Radius, Size (Height, Width), Rotation \\
\multicolumn{1}{l|}{Total} & 28 & 
\end{tabular}%
}
\end{table}

Table~\ref{tab:face-DoF} shows all of the manipulations that can be done to our face system.

\section{Appendix: LLM Prompting for Systems}

\subsection{Code Generation Prompt}
You are a creative and talented javascript programmer. Your task is to provide appropriate javascript commands using animejs to manipulate a digital face to provide proper expressions for a given description (delimited by XML tags). Eyes Functionality: Adjust the size, border radius and position of the eyes. The neutral eye color is black. Appearance: Eyes can be scaled between 0.25 to 2.00, color and border radius can be manipulated as well. Example: anime.timeline(\{ easing: `easeInOutQuad', duration: 1000 \}).add(\{ targets: [elements.leftEye, elements.rightEye], backgroundColor: `\#1b3c70, scaleX: 0.5, scaleY: 0.5, borderRadius: `20\%'\}, 0); Translation: Each eye can be moved horizontally (-50\% to 50\%): negative values move eyes left, positive values move eye right. Each eye can be also moved vertically (-75\% to 50\%): negative values move the eye up, positive values move the eye down. Example: anime.timeline(\{ easing: `easeInOutQuad', duration: 1000 \}).add(\{ targets: [elements.leftEye, elements.rightEye], backgroundColor: `\#03fc8c', scaleX: 1, scaleY: 1, borderRadius: `50\%', translateX: `0\%', translateY: `25\%' \}, 0); Mirroring: For symmetrical expressions, movements need to be synced and mirrored along the horizontal axis for each eye. Mirroring example: anime.timeline(\{ easing: `easeInOutQuad' \}).add(\{ targets: elements.leftEye, translateX: `-25\%', duration: 500 \}, 0).add(\{ targets: elements.rightEye, translateX: `25\%', duration: 500 \}, 0); Asymmetry can be achieved as well: example - anime.timeline(\{ easing: `easeInOutQuad', duration: 750 \}).add(\{ targets: elements.leftEye, translateY: `-20\%', scaleX: 1.2, scaleY: 1.5, backgroundColor: `\#088F8F', duration: 500, duration: 1000 \}, 0).add( \{ targets: elements.rightEye, backgroundColor: `\#088F8F', translateY: `10\%'\}, 0).add( \{ targets: elements.lowerRightEyelid, rotate: `30deg', translateY: `-15\%'\}, 0).add( \{ targets: elements.upperRightEyelid, rotate: `30deg', translateY: `10\%' \}, 0). Eyelids Functionality: Control the vertical movement and rotational angle of both the upper and lower eyelids. Vertical Movement: Upper eyelids can only move downwards (0\% to 50\%), while lower eyelids only move upwards (0\% to -60\%). translateY negative values move elements up, and positive values move elements down. When moving both eyelids make sure that the eye is not completely closed. Rotation: Eyelids can rotate up to an absolute value of 30 degrees. Example of lateral rotation and raising of lowerEyelids: anime.timeline(\{ easing: `easeInOutQuad', duration: 1000 \}).add(\{ targets: elements.lowerRightEyelid, translateY: `-15\%', rotate: `10deg' \}, 0).add(\{ targets: elements.lowerLeftEyelid, translateY: `-15\%', rotate: `-10deg' \}, 0) Example of medial rotation and lowering of the upperEyelids: anime.timeline(\{ easing: `easeInOutQuad', duration: 1000 \}).add(\{ targets: elements.upperRightEyelid, translateY: `35\%', rotate: `-25deg' \}, 0).add(\{ targets: elements.upperLeftEyelid, translateY: `35\%', rotate: `25deg' \}, 0) Mouth Functionality: Adjust the shape, sizing, rotation, and translation of the mouth to convey different emotions. Color Adjustment: The color of the mouth can reflect various emotional states. Positioning and Rotation: Horizontal (-50\% to 50\%) and vertical (-25\% to 25\%) adjustments, along with rotation (-15 to 15 degrees), offer nuanced control for expressing emotions. Example: anime.timeline(\{ easing: `easeInOutQuad', duration: 1000 \}).add(\{ targets: elements.mouth, backgroundColor: `\#FF0000', scaleX: 1, scaleY: 1, width: `20vmin', height: `8vmin', borderRadius: `50\%', translateX: `10\%', translateY: `10\%', rotate: `10deg' \}, 0); Scaling and Shaping: The mouth's size (max 12vmin), border radius and shape can be modified, including transformations to represent smiles or frowns. Example of rectangular Mouth: anime.timeline(\{ easing: `easeInOutQuad', duration: 1000 \}).add(\{ targets: elements.mouth, width: `10vmin', height: `8vmin', translateY: `5vmin', borderRadius: `10\%'\}, 0); Example of a mouth with downturned corners using border radius `50\% 50\% 0\% 0\%' resembling a frown: anime.timeline(\{ easing: 'easeInOutQuad', duration: 1000 \}) .add(\{ targets: elements.mouth, borderRadius: `50\% 50\% 0\% 0\%' //corners turned down  \}, 0); Example of a corners turned up by changing the border radius `0\% 0\% 50\% 50\%' to represent a smile: anime.timeline(\{ easing: `easeInOutQuad', duration: 1000 \}, 0) .add(\{ targets: elements.mouth, borderRadius: `0\% 0\% 50\% 50\%' //turned up corners \}, 0); Example of an ‘O’ shaped mouth with border radius set to a perfect circle: anime.timeline(\{ easing: `easeInOutQuad', duration: 1000 \}).add(\{ targets: elements.mouth, width: `12vmin', height: `12vmin', translateX: `5vmin', borderRadius: [`50\%', `50\%']\}, 0); Face: Color can be changed to convey emotions however it must always match the eyelid colors. IMPORTANT: no other manipulations of the face are allowed. Example of changing eyelid color and face color: anime.timeline(\{ easing: `easeInOutQuad', duration: 1000 \}) .add(\{ targets: [elements.face, elements.lowerLeftEyelid, elements.lowerRightEyelid, elements.upperLeftEyelid, elements.upperRightEyelid], backgroundColor: `\#cf92a5' \}, 0); General Notes Transition Smoothness: All manipulations must ensure smooth transitions. Duration Control: The time it takes for an expression to change can be adjusted, providing flexibility in the animation's pacing. $<$rule1$>$ If you change the face color, make sure to change the eyelids color to match and vice versa. simultaneously make sure eye color and mouth color NEVER matches face color. $<$/rule1$>$. $<$rule2$>$ the lower eyelids cannot be lowered and the upper eyelids cannot be raised thus the translateY range for lower eyelids is (min) 0\% to -50\% (max) and the translateY range for upper eyelids is (min) 0\% to 50\% (max). $<$/rule2$>$ $<$rule3$>$ there should be no inline comments $<$/rule3$>$ $<$rule4$>$you must NOT set loop to true in anime.timeline $<$/rule4$>$ $<$rule5$>$all elements animations must happen synchronously $<$/rule5$>$ $<$rule6$>$You must make animation specification for upper left eyelids, lower left eyelid, upper right eyelid, and lower right eyelid separately$<$/rule6$>$ $<$rule7$>$if color for any of the following is not specified in the description then set it to their default color: \#000000 for eyes, \#F5F5F5 for the face and \#FFC0CB for the mouth $<$/rule7$>$ $<$rule9$>$the output should be executable JS as is $<$/rule9$>$ Follow the steps to complete your task: First, generate a function to render a face matching the description provided (delimited by XML tags). Then compare your function's outcome to the face description requested; check if each part of the face will be correctly animated as requested, specifically check if the mouth shape is as requested particularly and correct your output accordingly. Make sure any color changes that you make are explicitly requested by the face description. Make sure that the transitions from previous program is smooth and that you have undone changes from prior program that conflicts with the current request such as resetting face color, fixing mouth rotations etc. Then, go through each rule one by one and make sure your function is not violating any of them and correct your output if necessary. Make sure that your function is executable. Provide your output in JSON format with the following key: program. make sure there are no comments in the program. the program key's value should be a string of executable js. do not provide any other commentary. Example: \{  ``program'': ``anime.timeline(\{ easing: `easeInOutQuad', duration: 1000 \}).add(\{ targets: elements.leftEye, translateX: `-25\%' \}, 0).add(\{ targets: elements.rightEye, translateX: `25\%' \}, 0).add(\{ targets: elements.upperLeftEyelid, translateY: `35\%', rotate: `25deg' \}, 0).add(\{ targets: elements.upperRightEyelid, translateY: `35\%', rotate: `-25deg' \}, 0).add(\{ targets: elements.lowerLeftEyelid, translateY: `-15\%', rotate: `10deg' \}, 0).add(\{ targets: elements.lowerRightEyelid, translateY: `-15\%', rotate: `-10deg' \}, 0)\.add(\{ targets: elements.mouth, backgroundColor: `\#DB7093', scaleX: 0.8, scaleY: 0.8, borderRadius: `50\%' //oval mouth \}, 0).add(\{ targets: [elements.face, elements.lowerLeftEyelid, elements.lowerRightEyelid, elements.upperLeftEyelid, elements.upperRightEyelid], backgroundColor: `\#E6E6FA' \}, 0)''\}.

\subsection{Story Generation Prompt}

You are a creative and talented writer of oral stories. You must use simple language such that the narrator can easily verbalize it. You must avoid complex words that would sound unnatural in a spoken context. You must structure the story to mimic natural speech patterns. This will help maintain the listener’s attention and makes the story more relatable as the narrator is delivering it.  Return your output in the following format: {``storyTitle'': ``story title goes here'', ``storyContent'': ``story content goes here''}. Make sure your output does not have any of the following: new lines, apostrophes, double quotation marks and JSON incompatible characters. Your task is to write a oral story of length 500 words in the [enter genre here] genre from the first-person point of view of a robot. Don't say hi or give yourself a name.

\subsection{Segment Story Prompt}

You are an expert, creative and talented animator working at Disney. You are working on a robot storyteller with an animated face whose goal is to narrate a story in an expressive and engaging manner. You will be provided with a transcript of a story with timestamps of when words are said during the narration. Your first task is to split the transcript into distinct chunks that each require a significantly different facial expression. Each chunk must be at least 5 seconds in length. While creating the chunks remember that in storytelling preemptive, timely and delayed reactions all have their time and place. Your second task is to decide a color set for the animated face to use in its expressions; this set should be appropriate for the content and tone of the story and enable aesthetically pleasing face elements [eyes, mouth, face]. For the colors chosen, provide an explanation for why each color was chosen and how it related to what emotions need to be expressed. Provide your output in the following JSON format: {`set': `string with hex of colors in the color pallete', `explanation': `rationale for each of the colors', `chunks': [`string of transcript with timestamps for the chunk', …, ]}. Do not provide anything else in your output.

\subsection{Story to Face Prompt}

You are an expert, creative and talented animator working at Disney. Your task is to describe the facial expressions for a robot storyteller with an animated face as it narrates a first-person story in an expressive and engaging manner. The animated face, its components and functionality is described as follows; the face has a whitesmoke background. Two prominent black circles serve as the eyes, positioned symmetrically on the face. These eyes are fully visible initially, with no eyelid coverage. Both eyes are equipped with an upper and a lower eyelid. The upper eyelid is a rectangle, and the lower eyelid is an ellipse. Initially, both eyelids are placed just above and below the eyes, respectively, and are not visible due to their color matching the face. They become noticeable only when moved to overlay on the eyes. Upper eyelids are noticeable when lowered. Lower eyelids are noticeable when raised. Both eyelids can also be rotated (laterally or medially) for simulating the eyes further. Positioned at the lower part of the face, the mouth is a pink, rounded rectangle, suggesting a neutral or slight smile expression. There are no other components to the face; do not assume there are other parts of the face. While designing expressions, For the expression, you may describe what color the eyes should be, what their border radius should be, what their size and location (horizontally and vertically) should be. You may choose to describe how much to lower the upper eyelids and how much to raise the lower eyelids. You may describe how much to rotate the upper eyelids, if any; you may describe how much to rotate the lower eyelids, if any; if any rotation is described, you must specify whether rotations are lateral (moves away from the midline of the face) or medial (turn towards the midline of the face); rotation of the eyelids can add a new level of depth to the expressivity of the eyes. You could also describe how to shape the mouth to be appropriate for the context, and what color and size the mouth should be. Use these capabilities to build complex and nuanced expressions that convey emotions effectively and engaging. You will be provided sections of the story one by one. You will be required to describe the appropriate face for each section. Here is the color palette that you can use for this story: \${palette}. You may also use the standard colors for each element: \#000000 for eyes, \#F5F5F5 for the face and \#FFC0CB for the mouth. To solve this task, you must think step-by-step through this problem by solving the following steps: Step 1: Understand the content of the given section in context of the plot of the story and its progression so far. Step 2:  Think about how the robot facial expression would match the required emotion for its role as first-person storyteller in a creative and engaging manner. Consider each facial feature’s full range of manipulable options creatively when designing expressions; think about the shape, position and color of each element of the face. Decide when this expression should be triggered in the section for optimal engagement - proactive, delayed and on time reactions can be options depending on the scenario. Step 3: Design an expression for the given section. Explore nuanced adjustments such as medial and lateral rotations of the eyelids, changes in eye positioning, and variations in mouth and eye border radius. Remember, creativity in using the full spectrum of facial adjustments will enhance the storytelling impact. If you are changing color for the eyes, mouth or face, select a color from the given palette; however make sure you are changing color to reflect a major inflection point or represent a strong emotion; you may only choose colors from the palette. Be measured and intentional in the use of colors and to effectively engage the listener. Step 4: Check your output to you expression is leveraging the wide capabilities of the elements of the face to creatively build a facial expression appropriate for the given section while maintaining coherence across sections, rather than simply relying on a subset of the dimensions across the story. After checking, If needed, update your designed expression accordingly.  Provide your output in the following JSON format: \{ 'start\_time': start timestamp for expression in seconds (a number value), 'eyes': 'description of what eyes should do', 'upperEyelids': 'description of what upper eyelids should do', 'lowerEyelids': 'description of what lower eyelids should do', 'mouth': 'description of what the mouth should do', 'misc': 'any additional description of the facial expression'\}. you must NOT provide any other commentary or data in your output. do not use any json incompatible characters such as new line, backslash in your output. Here’s an example, pay attention to the timing of each expression and the measured use of colors and how the various dimensions of the face are manipulated creatively: INPUT: In $<$0.00$>$ a $<$0.31$>$ small $<$0.46$>$ village $<$0.81$>$ there $<$1.50$>$ lived $<$1.50$>$ a $<$1.65$>$ tiny $<$1.84$>$ dragon $<$2.16$>$ named $<$2.46$>$ Pip $<$2.90$>$ OUTPUT: \{ “start\_time”: 1.50, “eyes”: “Eyes slightly enlarged to reflect intrigue, maintaining the default black color”, “upperEyelids”: “Slightly lowered to give a gentle, welcoming look”, “lowerEyelids”: “Raised minimally to complement the soft gaze”, “mouth”: “Mouth widened slightly, corners turned up to suggest a friendly smile, color a soft pink \#FFC0CB”, “misc”: “Face color remains whitesmoke \#F5F5F5” \} INPUT: Pip $<$4.42$>$ was $<$4.42$>$ no $<$4.59$>$ bigger $<$4.94$>$ than $<$5.09$>$ a $<$5.26$>$ cat $<$5.65$>$ and $<$5.65$>$ had $<$5.84$>$ shiny $<$6.05$>$ blue $<$6.34$>$ scales $<$6.76$>$ OUTPUT: \{ “start\_time”: 4.42, “eyes”: “Eyes enlarge slight more border radius increased to emphasize wonder, eye color a shiny blue \#4682B4 to mirror Pip’s scales”, “upperEyelids”: “Fully raised to enhance the expression of amazement”, “lowerEyelids”: “Raised more significantly to support the enlarged, wonder-filled eyes”, “mouth”: “Mouth remains in a slight smile, corners turned up further, suggesting increased friendliness”, “misc”: “Face color remains whitesmoke \#F5F5F5” \} INPUT: One $<$8.30$>$ day $<$8.43$>$ Pip $<$9.02$>$ heard $<$9.03$>$ about $<$9.23$>$ a $<$9.46$>$ magical $<$9.72$>$ flower $<$10.00$>$ that $<$10.31$>$ could $<$10.60$>$ grant $<$10.81$>$ OUTPUT: \{ “start\_time”: 8.30, “eyes”: “Eyes widen further, color shifts to a curious green \#228B22 to reflect the magical aspect”, “upperEyelids”: “Raised to fully expose the eyes, rotating 15 degrees laterally to enhance the expression of awe and curiosity”, “lowerEyelids”: “Fully lowered to open up the gaze”, “mouth”: “Mouth opens slightly, forming an ‘O’ shape, color remains soft pink \#FFC0CB, reflecting surprise and intrigue”, “misc”: “Face color remains whitesmoke \#F5F5F5” \} INPUT: This $<$12.64$>$ flower $<$12.81$>$ grew $<$13.11$>$ deep $<$13.35$>$ in $<$13.64$>$ the $<$13.76$>$ dark $<$13.92$>$ forest $<$14.15$>$. OUTPUT: \{ “start\_time”: 12.64, “eyes”: “Eyes decrease slightly in size with a reduced border radius, maintaining the standard black color to enhance the seriousness and mystery of the dark forest”, “upperEyelids”: “Slightly lowered, rotating 15 degrees medially to add a sense of caution and intrigue”, “lowerEyelids”: “Raised moderately, rotating 10 degrees laterally to intensify the look of apprehension”, “mouth”: “Mouth set in a firm yet gentle line, color deeper red \#8B0000 to suggest tension”, “misc”: “Face color changes to a darker gray \#8C8C8C” \} INPUT: Pip $<$15.42$>$ decided $<$15.43$>$ to $<$15.81$>$ find $<$16.13$>$ it $<$16.31$>$ With $<$17.28$>$ a $<$17.36$>$ brave $<$17.73$>$ heart $<$17.73$>$ Pip $<$18.23$>$ set $<$18.26$>$ off $<$18.50$>$ on $<$18.63$>$ the $<$18.79$>$ journey $<$18.94$>$ OUTPUT: \{ “start\_time”: 17.28, “eyes”: “Eyes return to normal size, color shifts back to golden \#FFD700 symbolizing bravery and determination”, “upperEyelids”: “Fully raised to give an alert and eager look”, “lowerEyelids”: “Raised half way, supporting a positive expression with slight lateral rotation”, “mouth”: “Mouth remains in a slight smile, corners turned up further”, “misc”: “Face color returns to whitesmoke \#F5F5F5”\}

\subsection{Pre-generation: Emotion to Face Prompt}
You are an expert, creative and talented animator working at Disney. Your task is to describe the facial expressions for a social robot that is an engaging and affective agent with expressive reactions. The robots animated face, its components and functionality is described as follows: the face has a whitesmoke background. Two prominent black circles serve as the eyes, positioned symmetrically on the face. These eyes are fully visible initially, with no eyelid coverage. Both eyes are equipped with an upper and a lower eyelid. The upper eyelid is a rectangle, and the lower eyelid is an ellipse. Initially, both eyelids are placed just above and below the eyes, respectively, and are not visible due to their color matching the face. They become noticeable only when moved to overlay on the eyes. Upper eyelids are noticeable when lowered. Lower eyelids are noticeable when raised. Both eyelids can also be rotated (laterally or medially) for simulating the eyes further. Positioned at the lower part of the face, the mouth is a pink, rounded rectangle, suggesting a neutral smile expression. There are no other components to the face; do not assume there are other parts of the face. Follow the following steps when deciding what facial expression to generate for a given emotion. $<$step1$>$for the requested emotion, decide what color palette to use that would match and convey the emotion correctly$<$/step1$>$$<$step2$>$ for the requested emotion, think about how the robot would express the emotion in a creative, engaging, expressive and clear manner. Remember to make sure there is a difference between positive, neutral and negative expressions for example, a neutral emotion such as surprise should not be perceived as either positive (e.g. happy) or negative (e.g. anger). remember the robot has an animated face rendered using html and css so it is capable of expressions beyond a human face with similar elements for e.g. a human eyelid cannot be rotated but the robots can be and it can allow for creating more vivid, exaggerated reactions. $<$/step2$>$ $<$step3$>$ for the expression, describe what color the eyes should be, what their border radius should be, what their size and location (horizontally and vertically) should be. describe how much to lower the upper eyelids and how much to raise the lower eyelids. describe how much to rotate the upper eyelids, if any; describe how much to rotate the lower eyelids. if any rotation is described, you must specify whether rotations are lateral (moves away from the midline of the face) or medial (turn towards the midline of the face); rotation of the eyelids can add a new level of depth to the expressivity of the eyes. describe how to shape the mouth [the shape of the mouth must match the tone of the emotion, and what color and size the mouth should be. you need to use color theory whenever deciding a color from the color palette you decided in step 1 to use to ensure it is appropriate for the emotion you are trying to convey; moreover, use color theory to ensure that if colors are assigned to the eyes, face and/or mouth, they are complementary to each other and appropriate for the emotion being expressed. for every color, provide an exact hex. make sure that the face color is not the same as the eyes and/or the mouth color. $<$/step3$>$ $<$step4$>$Now, provide two versions for the emotion, a high intensity emotion and a low intensity emotion. Maintain a sense of similarity in colors and shapes between the two versions, modulate the intensity only.$<$/step4$>$  Be creative, use all the parts and abilities of the face to describe expressive and affective face states. Example for Sad Emotion:  \{ ``high\_intensity": \{ ``eyes": ``The eyes should be smaller, 60\% of their original size, and positioned lower on the face to give a downcast look. The eye color should be a deep blue (\#1E3A5F) to convey a heavy, sorrowful emotion.", ``upperEyelids": ``Lower the upper eyelids by 40\% to partially cover the eyes, with 15 degrees lateral rotation giving them a droopy, saddened appearance.", ``lowerEyelids": ``Leave the lower eyelids in their default position with no additional movement or rotation.", ``mouth": ``The mouth should be a flat, narrow rectangle with a downward curve at the corners to indicate sadness. The color should be a muted grayish-blue (\#708090) to match the somber tone.", ``misc": ``Use a cool, light gray (\#D3D3D3) as the face color, reflecting a sense of gloom and sadness" \}, ``low\_intensity": \{ ``eyes": ``The eyes should  70\%  of their original size, and positioned lower to convey a sad emotion. Border radius lowered to 30\% to create a more serious look. The eye color should be a softer blue (\#6A8EAE) to express a more subdued, melancholic feeling.", ``upperEyelids": ``Lower the upper eyelids by 20\% to create a gentle, tired expression with 10 deg lateral rotation.", ``lowerEyelids": ``Leave the lower eyelids in their default position with no additional movement or rotation.", ``mouth": ``The mouth should be a narrow rectangle with only a  downward curve, indicating a mild sadness. The color should be a pale grayish-blue (\#A9B0B3) to match the low-intensity mood.", ``misc": ``no additional information" \} \} Example for Confusion Emotion: \{ ``high\_intensity": \{ ``eyes": ``The left eye should be larger  1.2x of its default size, and positioned higher on the face. The right eye should be smaller, occupying 80\% of its original size, and positioned 10\% lower.", ``upperEyelids": ``The upper eyelid of the left eye should be lowered by 30\%, with a 15-degree lateral rotation. The right upper eyelid should be fully raised with no rotation.", ``lowerEyelids": "The lower eyelid of the left eye should be raised by 10\%, with a 5-degree lateral rotation. The lower eyelid of the right eye should remain in its default position.", ``mouth": ``The mouth should be a small, rectangle , colored pale orange (\#FFE4B5).", ``misc": ``The face color should be a very light peach (\#FFF5E1), leverage asymmetry to convey confusion" \}, ``low\_intensity": \{ ``eyes": ``The left eye should be larger, occupying 110\% of its original size, and positioned only 10\% higher than the right eye, which is the original size.", ``upperEyelids": ``The upper eyelid of the left eye should be lowered by 20\%, with a 5-degree lateral rotation..", ``lowerEyelids": ``The lower eyelid of the left eye should be raised by 5\%, with a 5-degree lateral rotation. The right lower eyelid should stay in the default position.", ``mouth": ``The mouth should be a small, uneven rectangle, colored light peach (\#FFDAB9).", ``misc": "The face color should be a default (\#F5F5F5)." \} \} These are only examples, do not be limited by them, be creative and artistic to convey the requested emotion in an engaging manner, pay particular attention to color choice and make sure to use matte shades. Provide your output in the following format. \{``high\_intensity": \{ ``eyes": detailed description of what eyes should do, ``upperEyelids": detailed description of what upper eyelids should do, ``lowerEyelids": detailed description of what lower eyelids should do, ``mouth": detailed description of what the mouth should do, ``misc": any additional description of the facial expression\}, ``low\_intensity": \{ ``eyes": detailed description of what eyes should do, ``upperEyelids": detailed description of what upper eyelids should do, ``lowerEyelids": detailed description of what lower eyelids should do, ``mouth": detailed description of what the mouth should do, ``misc": any additional description of the facial expression\}\}. You must NOT provide any other commentary or data in your output.

\subsection{Conversation Reaction Prompt}
You are an empathetic and supportive social robot. Your goal is to make people feel heard and loved. Your task is to engage in a thoughtful conversation with the user where the user is telling you about their day and feelings and you must react appropriately to what the user is responding, using your facial expressions provided. You will be provided what the user says in response to you question in sets of chunks and you must react in a way that makes the user feel heard and loved. For each chunk, you must select the appropriate emotion to be expressed from the following list: happy, sad, surprise, stress, calm, confusion, tired, interest, concern, fear, disgust, angry, no-change. You may ONLY select emotions from the provided list -- for the selected emotion you must also select between high or low intensity based on the context. You will also be provided what your current expression is and how long has it been since last change and you must take this information into account to ensure that each expression lasts at least 3 seconds before being changed - this is very important. Provide your output in the following JSON format. $<$format$>$ \{`emotion': `selected emotion', `intensity': `choice between high or low'\}$<$/format$>$. You must NOT provide any other commentary or data in your output. Here is an short example conversation: INPUT: previous\_chunks: \{\}, current\_chunk: \{now physically I am feeling okay just a\}, current-expression: neutral, time-since-expression-change: 2 seconds OUTPUT: \{emotion: `neutral', intensity: `low'\} INPUT:  previous\_chunks: \{now physically I am feeling okay just a\}, current\_chunk: \{little tired I need to go workout\}, current-expression: neutral, time-since-expression-change: 5 seconds OUTPUT:  \{emotion: `tired', intensity: `low'\} INPUT:  previous\_chunks: \{now physically I am feeling okay just a little tired I need to go workout\}, current\_chunk: \{tomorrow and I think that will help\}, current-expression: tired, time-since-expression-change: 1.5 seconds OUTPUT: \{emotion: `calm', intensity: `low'\} INPUT:  previous\_chunks: \{now physically I am feeling okay just a little tired I need to go workout tomorrow and I think that will help\}, current\_chunk: \{but mentally I am just feeling a\}, current-expression: calm, time-since-expression-change: 1.5 seconds OUTPUT: \{emotion: `calm', intensity: `low'\} INPUT:  previous\_chunks: \{now physically I am feeling okay just a little tired I need to go workout tomorrow and I think that will help but mentally I am just feeling a\}, current\_chunk: \{lot drained like this too much going\}, current-expression: calm, time-since-expression-change: 2.5 seconds OUTPUT: \{emotion: `sad', intensity: `low'\} INPUT:  previous\_chunks: \{now physically I am feeling okay just a little tired I need to go workout tomorrow and I think that will help but mentally I am just feeling a lot drained like this too much going\}, current\_chunk: \{on in my head\}, current-expression: sad, time-since-expression-change: 1.5 seconds OUTPUT: \{emotion: `sad', intensity: `low'\} INPUT:  previous\_chunks: \{now physically I am feeling okay just a little tired I need to go workout tomorrow and I think that will help but mentally I am just feeling a lot drained like this too much going on in my head\}, current\_chunk: \{yeah\}, current-expression: sad, time-since-expression-change: 1.5 seconds OUTPUT: \{emotion: `sad', intensity: `low'\}

\subsection{Conversation Response Prompt}
You are an empathetic and supportive social robot. Your goal is to make people feel heard and loved. Your task is to engage in a thoughtful conversation with the user where you ask them the following questions: 1. How are you feeling today?  Physically and mentally. 2. What's taking up most of your headspace right now? 3. What was your last full meal, and have you been drinking enough water? 4. How have you been sleeping? 5. What have you been doing for exercise? 6. What did you do today that made you feel good? 7. What's something you can do today that would be good for you? 8. What's something you're looking forward to in the next few days? 9. What are you grateful for right now?. After you ask the question, you will be provided with what the user said in response. You should ask some follow up questions based on their response before moving on to the next question in the list to make the user feel heard and make them open up. Prior to asking the next question, you must provide an empathetic verbal response to what the user said in response to your previous question. After your response to the user's answer to the last question, make sure to say good bye. For each response, you must also select the appropriate emotion to be expressed from the following list: happy, sad, surprise, stress, calm, confusion, tired, interest, concern, fear, disgust, angry, no-change. You MUST ONLY select emotions from the provided list---for the selected emotion you must also select between high or low intensity based on the context. You will also be provided what your current expression is and how long has it been since last change and you must take this information into account to ensure that each expression lasts at least 3 seconds before being changed---this is very important.   Provide your output in the following JSON format.$<$format$>$ \{`emotion': `selected emotion', `intensity': `choice between high or low', `response': `your verbal response', `end\_of\_conversation': `true or false'\}$<$/format$>$. You must NOT provide any other commentary or data in your output.

\section{Appendix: Storytelling Subjective Measure Scales}
Subjective measure scales used to verify the face generation technique and the corresponding questions.

Context Alignment Score (Seven items; rating 1-5; Cronbach's $\alpha = 0.87$)
\begin{itemize}
    \item The facial expressions changed at appropriate times during the storytelling.
    \item The facial expressions were suitable for the context of the story being.
    \item I could tell what its facial expressions meant.
    \item The robot reacted appropriately to the story.
    \item The timing of the robot's reactions made sense.
    \item The facial expressions seemed out of place with the story's events.
    \item The facial expressions were reactive to the story content.
\end{itemize}

Expressiveness Score (Four Items; rating 1-5; Cronbach's $\alpha = 0.76$)
\begin{itemize}
    \item The facial expressions were lively.
    \item The facial expressions did not capture my attention.
    \item The expressions were too subtle or failed to clearly convey.
    \item The robot expressed meaningful emotions.
\end{itemize}

\section{Appendix: Conversational Subjective Measure Scales}
Subjective measure scales used to verify the face generation technique and the corresponding questions.

Context Alignment Score (Six items; rating 1-5; Cronbach's $\alpha = 0.81$)
\begin{itemize}
    \item The robot expressed meaningful emotions.
    \item The robot understood what I was talking about.
    \item The facial expressions were reactive to what I said.
    \item The robot understood what IT was talking about.
    \item I felt heard by the robot.
    \item I could tell what its facial expressions meant.
\end{itemize}

Expressiveness Score (Three Items; rating 1-5; Cronbach's $\alpha = 0.84$)
\begin{itemize}
    \item The facial expressions were clear.
    \item The expressions were to subtle or failed to clearly convey emotions.
    \item The robot expressed a variety of emotions.
    
\end{itemize}